\documentclass[runningheads]{llncs}

\usepackage[T1]{fontenc}
\usepackage{graphicx}
\usepackage{amsmath}
\usepackage{amssymb}

\usepackage{microtype} 
\usepackage{booktabs}

\begin{document}

\title{Perspective Latents as an Architectural Condition for Causal Emergence in Active Inference Agents}
\titlerunning{Perspective Latents as an Architectural Condition for Causal Emergence}

\author{Hongju Pae\orcidID{0000-0002-5174-8858}}

\institute{Active Inference Institute, Crescent City, CA, USA\\
\email{hjpae@activeinference.institute}}


\maketitle

\begin{abstract}
A recent line of work measures causal emergence in reinforcement learning agents through Integrated Information Decomposition, reporting that $\Phi_r$ grows with training and tracks reward improvement. For active inference, this raises the question of how reward-free predictive organization relates to such information-theoretic signatures. I test this within an active inference agent whose architecture separates a fast perception latent $z$ from a slow global latent $g$, where $g$ is driven by prediction error and structurally decoupled from policy gradients. In a reward-free environmental regime-switching protocol, $\Phi_r$ concentrates in $g$; its aggregate magnitude is largely architectural and decreases with training. The substantive effect of learning becomes legible only at the atom-compositional level: decoupling flips sign from negative to positive and becomes regime-invariant under environmental change, while downward causation carries the regime-dependent adjustment. These results identify $g$ as the architectural locus of $\Phi_r$-relevant temporal organization in an active inference agent, and argue against reading scalar $\Phi_r$ as a direct index of learned integration.
\end{abstract}

\section{Introduction} \label{section1}

\par Recent work has used Integrated Information Decomposition ($\Phi$ID)~\cite{mediano2019phiid,mediano2025taxonomy}, an extension of the integrated information tradition~\cite{tononi2008integrated,albantakis2023iit} into the causal emergence framework~\cite{hoel2013emergence,rosas2020}, to analyze latent dynamics in trained reinforcement learning agents, reporting that the slow integrative organization measured by $\Phi_r$ grows with training and aligns globally with reward improvement~\cite{pigozzi2026causal}. This positions $\Phi_r$ as a candidate empirical signature of agent \textit{integration}: the extent to which an agent's internal state predicts its own future as a coherent whole, rather than as a collection of independent parts. Yet such analysis detects the slow integrative mode without specifying its architectural locus. Hence I ask: \textit{what architectural locus, within a single agent, gives rise to the slow integrative mode that $\Phi_r$ detects?}

\par A natural candidate is suggested by an independent line of work on \textit{perspective latents} in active inference agents. Phenomenological accounts of subjectivity have long emphasized that any subjective experience is oriented from a perspectival standpoint~\cite{husserl2001,merleau2013,zahavi2005}. This could be interpreted as a slowly evolving, history-sensitive condition under which the world becomes given as meaningful, threatening, or unremarkable. Recent active inference work has formalized analogous structures as slow latent variables that anchor an agent's interpretive stance toward the world across time. The architecture I adopt here introduces a slow latent $g$, driven by environmental prediction error minimization and decoupled from policy gradients, and demonstrates that $g$ exhibits hysteresis under regime change and reorganizes perceptual encoding as a function of accumulated history~\cite{pae2026aaai,pae2026sameworld,pae2026body}. The present paper asks whether $g$ is the locus that $\Phi_r$ detects. Rather than re-evaluating the behavioral phenotype of this agent, the present paper treats the inherited architecture as a fixed test case and asks where $\Phi_r$-relevant temporal organization appears within it.

\par This is tested in a reward-free regime-switching protocol, computing $\Phi_r$ over the fast perceptual latent $z$ and the slow perspective latent $g$ across 30 seeds. Three controls anchor the analysis: (1) a within-episode temporal shuffle (isolating temporal structure), (2) an untrained-architecture baseline (isolating learning), and (3) a pre/post regime-switch comparison (testing atom-level adaptivity). To characterize the locus of integration further, $\Phi_r$ is also decomposed into three theory-grounded atom groups, following \cite{rosas2020}: (i) decoupling, (ii) downward causation, and (iii) part-driven contributions.

\par On the basis of the perspective framework, three predictions follow. First, since $g$ is a history-sensitive Gated Recurrent Unit (GRU)-gated latent in the architecture, $\Phi_r$ should concentrate on $g$ rather than on $z$. Second, since $\Phi_r$ has been reported to grow with training under reward-driven learning~\cite{pigozzi2026causal}, a similar growth might be expected under the reward-free prediction-error objective used here, with trained $\Phi_r(g)$ exceeding its untrained baseline. Third, environmental regime shift should leave a measurable signature in the atom-level composition of $\Phi_r(g)$, consistent with the history sensitivity previously reported for $g$~\cite{pae2026aaai}. The first and third predictions are confirmed by the analyses below. The second yields a more informative reversal: $\Phi_r(g)$ does not simply grow with training, while the substantive effect of learning becomes visible only when $\Phi_r$ is decomposed into its constituent atom groups. Disentangling these atom groups reveals a structural transformation that scalar $\Phi_r$ alone cannot capture.

\paragraph{Key Contributions.} \textbf{(1) Empirical locus.} $\Phi_r$ in a perspective architecture concentrates in the slow latent $g$ rather than the fast latent $z$. \textbf{(2) Architecture vs.\ learning.} Aggregate $\Phi_r(g)$ is largely supplied by the recurrent architecture itself; learning lowers this magnitude. \textbf{(3) Atom-level dissociation.} Learning shifts decoupling from negative to positive and makes it approximately regime-invariant, while downward causation remains regime-adaptive.

\section{Experiment Methods} \label{section2}

\subsection{Agent Architecture and Simulation Environment}
\par The agent implementation and its simulation environment are inherited directly from~\cite{pae2026aaai}, with no additions beyond the analysis-side knobs. In this agent, the perspective layer and the action policy layer are designed to be structurally separated, where the perspective layer answers to ``What kind of world do I believe I am still in?'', while the policy layer answers to ``What should I do right now?''~\cite{pae2026aaai}.

\begin{figure}[t]
\centering
\includegraphics[width=1.0\textwidth]{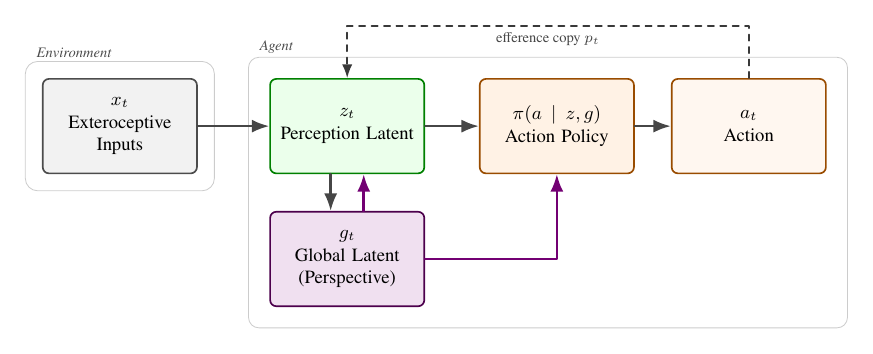}
\caption{\textbf{Agent architecture overview (inherited from~\cite{pae2026aaai}).} The fast perception latent $z_t$ and the slow global ``perspective'' latent $g_t$ feed into the action policy $\pi$, while $g_t$ is structurally decoupled from policy gradients. The present study analyzes the $\Phi_r$ structure of these latent trajectories.}
\label{fig1}
\end{figure}

\par Following the active inference principle, the agent learns to minimize its environmental prediction error. Agents act in a reward-free 2D gridworld of size $15 \times 9$ with 3 distinct environmental niche-like zones distinguished by zone-specific observation noise of $(\sigma_0, \sigma_1, \sigma_2 = 0.6, 0.3, 0.05)$. Environmental prediction is therefore harder in the leftmost zone and easier in the rightmost, so the agent comes to prefer rightward over leftward motion. Importantly, no reward signal is provided; the primary representational signal is one-step prediction error on perceptual reconstruction. This aligns with the active inference principle at the level of perception and world-modeling, while policy selection is trained through a learned actor loss rather than via expected-free-energy minimization; however, note that the paper's claims concern the perception and world-modeling side of the architecture rather than action-selection dynamics.

\par The observation encoder maps the agent's local surroundings (the 8 neighboring tiles in the 2D gridworld), together with the previous action encoded as a learned embedding $p_t \in \mathbb{R}^{8}$ (over the 5 actions UP / DOWN / LEFT / RIGHT / STAY), into a fast latent $z_t$. Architecturally, what matters is the separation of $z_t$, which serves the role of immediate perception, from $g_t$, which serves as a global-scale ``perspective'' that registers and predicts the environment at the regime level. It accumulates history through a damped GRU module updated from $(z_t,p_t)$; the parameters of this pathway are trained through the one-step prediction objective and the $g_t$-smoothness regularizer. A state head composes $z_t$, $p_t$ and $g_t$ into a policy state $s_t$, which feeds a discrete action policy $\pi(a_t \mid s_t)$ over the 5 actions. In all experiments, the dimensions are fixed at $z_t \in \mathbb{R}^{16}$, $p_t \in \mathbb{R}^{8}$, $g_t \in \mathbb{R}^{12}$, and $s_t \in \mathbb{R}^{16}$. An overview of the architecture is shown in Fig.~\ref{fig1}.

\par Importantly, $g_t$ is structurally decoupled from policy gradients through stop-gradient operators that sever the paths through which policy or actor losses could flow back into $g_t$. As a result, $g_t$ is largely shaped by global predictive coherence, rather than by actor-side gradients.

\subsection{Training Protocol}
\par Each seed is trained for $48,000$ environment steps with \texttt{Adam} optimizer at learning rate $3 \times 10^{-4}$ on $240$-step episodes; this corresponds to $200$ episodes per seed, with the network weights including those producing $z_t$ and $g_t$ carried across episode boundaries. A total of 30 seeds were run. 

\par The training objective (loss function) follows~\cite{pae2026aaai} and combines four terms:
\begin{equation}
\mathcal{L} = \mathcal{L}_{\text{pred}} + w_{\text{smooth}}\,\mathcal{L}_{\text{smooth}} + w_{\text{actor}}\,\mathcal{L}_{\text{actor}} - w_{\text{entropy}}\,\mathcal{H}(\pi)
\end{equation}
where $\mathcal{L}_{\text{pred}}$ is the one-step prediction error on the perceptual reconstruction (primary signal), $\mathcal{L}_{\text{smooth}}$ is a temporal smoothness regularizer on $g_t$, $\mathcal{L}_{\text{actor}}$ is an actor-style consistency cost applied through an exponential moving baseline with coefficient $\beta = 0.98$, and $\mathcal{H}(\pi)$ is the policy entropy. Loss weights are inherited unchanged from~\cite{pae2026aaai}: $w_{\text{smooth}} = 0.25$, $w_{\text{actor}} = 0.25$, $w_{\text{entropy}} = 0.001$. The actor term is gated on only after a $12,000$-step warmup, allowing the predictive backbone and $g$-dynamics to stabilize before actor-side optimization is introduced. Actor-side gradients are blocked from $g_t$ through the stop-gradient operators, so that the actor objective shapes $\pi$ and $s_t$ but not $g_t$ itself.

\subsection{Analysis Protocol}
\par For analysis, the trained weights are used to record replay rollouts under the same environment. Each replay episode runs for $T = 500$ steps, and the first $80$ steps are discarded as warmup. The architecture itself is identical between training and replay; only the regime-switch instrumentation and the episode length ($T = 500$ vs.\ $T = 240$) differ.

\paragraph{Replay conditions.} Two conditions are analyzed: a \textbf{\textit{control}} condition, in which the zone variances are fixed throughout at the same values used during training, and a \textbf{\textit{regime switch}} condition, in which the observation noise of the environment zones are reversed from $(\sigma_0, \sigma_1, \sigma_2) = (0.6, 0.3, 0.05)$ to $(0.05, 0.3, 0.6)$ at $t = 320$, defining a pre-switch window of $80 < t \leq 320$ and a post-switch window of $t > 320$. 10 rollout episodes per seed are collected per condition under stochastic action sampling, yielding $n = 300$ episodes per condition across $30$ seeds. For regime-switch analyses, $\Phi_r(g)$ and its atom-group decomposition are computed separately on the pre-switch window and the post-switch window. 


\paragraph{Control analyses.} Two additional controls separate contributions to the observed signals. A \textbf{\textit{within-episode temporal shuffle}} randomly permutes the time index of $g_t$ within each episode (three independent shuffles averaged per episode), preserving its marginal distribution while destroying all temporal order. This separates the contribution of temporal structure from that of the marginal distribution. 

\par An \textbf{\textit{untrained-architecture baseline}} instantiates the same architecture with the same hyperparameters but with fresh random initialization (random seeds are distinct from those used for trained checkpoints) and no training, with the replay protocol identical to the trained case. This separates the contribution of learning from that of the architecture alone.

\paragraph{Statistical analysis.} Paired comparisons use Student's paired $t$-test on episode-level estimates or within-episode pre/post regime switch differences, and unpaired group comparisons use Welch's unequal-variance $t$-test. Effect sizes are reported as Cohen's $d$ where informative. All $p$-values are two-sided, with significance assessed at $\alpha = 0.05$. Code and data are available at \url{https://github.com/hjpae/perspective-causal-emergence}.

\subsection{$\Phi_r$ Calculation}

\par To measure the temporal information structure of latent trajectories, I calculate $\Phi_r$, which is an estimator based on $\Phi$ID~\cite{mediano2019phiid,mediano2025taxonomy}, adapted from the implementation used in~\cite{pigozzi2026causal}. This estimator is local and pointwise, i.e. atom values are computed at each time step, are permitted to take negative values, and are summarized by the median over time. It therefore differs from the non-negative synergistic-channel construction that~\cite{rosas2020} uses to define causal decoupling as a scalar system-level property. In what follows, I borrow the atom taxonomy of~\cite{rosas2020} but not its non-negativity constraint. Thus, sign shifts in this study should be read as changes in the local pointwise contribution of the corresponding atom, rather than as changes in a Rosas-style non-negative index. The purpose of this analysis is to ask where $\Phi_r$-relevant temporal organization appears within the architecture, and how its composition changes with learning and regime shift. 


\par For a multivariate latent trajectory $X = [x_1,\dots,x_T]$, with $x_t \in \mathbb{R}^d$, the trajectory is first arranged as a $d \times T$ matrix and standardized dimension-wise using a corrected z-score procedure, which injects small noise into near-constant units before standardization. A lag-1 Gaussian mutual information matrix $M$ is then computed across latent dimensions, following the Gaussian copula approach to mutual information estimation~\cite{ince2017gcmi}. For each pair of dimensions, the lagged correlation $r_{ij}$ is converted to Gaussian mutual information as:
\begin{equation}
M_{ij} =
-\frac{1}{2}\log(1-r_{ij}^{2}),
\quad
M_{ii}=0
\label{eq2}
\end{equation}
Pairwise dependencies are retained only when significant under Bonferroni-corrected testing at $\alpha = 0.05$, and non-significant entries are set to zero. This matrix defines a weighted mutual information graph over latent dimensions.


\par The thresholded $M$ is then used to bisect the system. Except where explicitly noted below, the Fiedler vector of the graph Laplacian associated with $M$~\cite{fiedler1973} provides an automatic minimum information bipartition $min(A|B)$ of the $d$ dimensions; if the Fiedler split is degenerate, the implementation falls back to an index-based halving. The standardized trajectory $\tilde X$ is then collapsed onto a two-node trajectory $Y = (Y_A, Y_B)$ by within-partition averaging, determined as:
\begin{equation}
min(A|B) = \mathrm{Fiedler}(L(M)); \quad Y_A(t) = \frac{1}{|A|}\sum_{i\in A} \tilde X_i(t), \quad Y_B(t) = \frac{1}{|B|}\sum_{i\in B} \tilde X_i(t)
\label{eq3}
\end{equation}

\par This two-node trajectory $Y$ is the input to $\Phi$ID itself. A local $\Phi$ID lattice $\Pi(Y)$ is computed using Gaussian local entropy estimates and M\"obius inversion, producing local atom values $\pi_a(t)$ for each lattice atom $a$. Let $\mathcal{A}_r$ denote the subset of lattice atoms that contribute to $\Phi_r$. The episode-level $\Phi_r(X)$ is the median over time of the local sum, which could be expressed as:
\begin{equation}
\pi_a(t) = [\Pi(Y)]_a(t); \qquad \phi_r(t) = \sum_{a \in \mathcal{A}_r} \pi_a(t), \quad \Phi_r(X) = \mathrm{median}_t\,\phi_r(t)
\label{eq4}
\end{equation}

\par Throughout the results, I refer to this episode-level scalar $\Phi_r(X)$ as \textbf{\textit{aggregate} $\Phi_r$}. Here, ``aggregate'' means that the nine $\Phi_r$-relevant atom values have been summed locally and then summarized over time by the median. It does not refer to averaging across episodes or seeds, which are reported separately as descriptive statistics. Atom grouping is further discussed below. 

\par In the main analyses, I compute $\Phi_r$ separately for $z$ and $g$. For these $z$-only and $g$-only analyses, the bipartition is selected automatically by the Fiedler procedure described above. I also compute a diagnostic $\Phi_r$ on the joint trajectory $[z,g]$, forcing the bipartition to respect the architectural boundary $z|g$. This forced split asks how much temporal organization is present across the perceptual-perspective boundary, rather than how the Fiedler procedure would partition the concatenated latent space. Because the central signal localizes to $g$, the main results focus on the $g$ trajectory and its atom-level composition.

\paragraph{Atom grouping.}

\par To interpret the composition of $\Phi_r$, I decompose each $g$-trajectory into the nine $\Phi_r$-relevant $\Phi$ID atoms and group them by their directional pattern on the two-node partial information decomposition (PID) lattice~\cite{williams2010pid}. The nine atoms are those on the double-redundancy lattice whose source and target both involve the two-node partition induced by the Fiedler bisection; the atom taxonomy is that of~\cite{rosas2020}, and the specific implementation of the pointwise decomposition follows~\cite{pigozzi2026causal}. This grouping preserves the directional distinction of~\cite{rosas2020}: decoupling and downward terms correspond most directly to emergence-related whole-to-whole and whole-to-part organization, whereas the part-driven group closes the accounting of the full $\Phi_r$ sum in the present estimator. Because the bipartitions are produced by the Fiedler bisection, the labels of the individual parts might possess redundancy. Hence, I report 3 topology-level group sums rather than treating the 9 individual atom labels.

\par The first group is \textbf{\textit{decoupling}} (\textsc{whole} $\to$ \textsc{whole}), consisting of the single atom in which the whole at the present informs the whole at the future. This is the atom most directly associated with whole-as-whole irreducibility. The second group is \textbf{\textit{downward causation}} (\textsc{whole} $\to$ \textsc{part}), consisting of the 3 atoms in which whole-level present information contributes to future part-level structure. The third group is \textbf{\textit{part-driven contribution}} (\textsc{part} $\to$ \textsc{whole} and \textsc{part} $\to$ \textsc{part}), consisting of the remaining 5 $\Phi_r$-relevant atoms, including bottom-up and lateral part-mediated terms.

\section{Results} \label{section3}
\par The results are organized around the progression from localization to interpretation. First, I ask where $\Phi_r$-relevant temporal structure appears within the inherited architecture. Second, I test whether aggregate $\Phi_r(g)$ is produced by learning or already supplied by the recurrent substrate. Third, I decompose $\Phi_r(g)$ into atom groups to determine what learning changes. Finally, I examine whether the learned composition of $\Phi_r(g)$ responds differently to environmental regime shift. Taken together, these analyses position the perspective latent $g$ as the locus at which $\Phi_r$-relevant temporal organization can be decomposed and interpreted.

\subsection{$\Phi_r$ localizes to the perspective $g$, and depends on temporal order} 

\begin{figure}[t]
\centering
\includegraphics[width=1.0\textwidth]{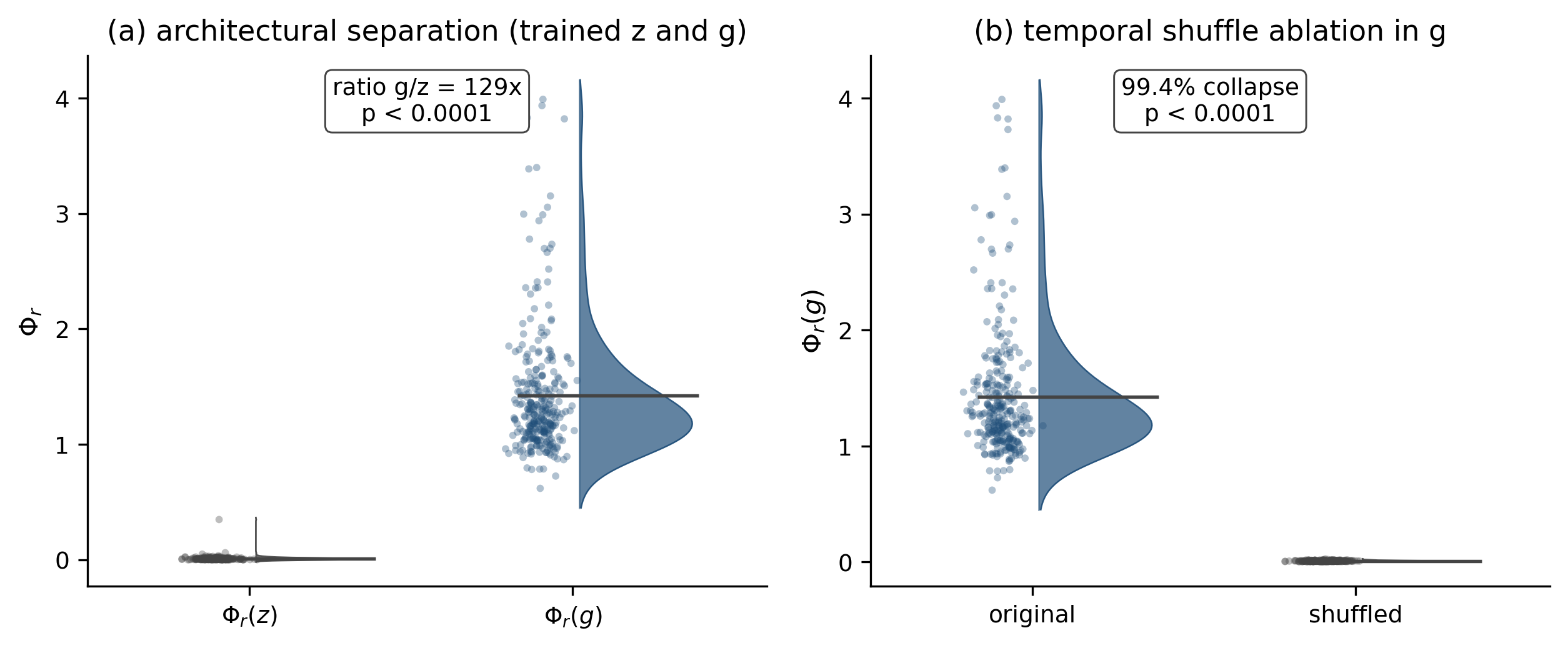}
\caption{\textbf{$\Phi_r$ localizes to the perspective latent and depends on temporal order.} Scatter points show individual episodes, the half-violin shows the pooled distribution, and the horizontal bar marks the mean. \textbf{(a).} Under the same replay condition, $\Phi_r(g)$ is $129\times$ larger than $\Phi_r(z)$ (mean $\Phi_r(g)=1.42$ $\pm$ SD $0.56$, mean $\Phi_r(z)=0.011$ $\pm$ SD $0.021$). \textbf{(b).} Within-episode temporal shuffling of $g$ collapses $\Phi_r(g)$ from $1.42 \pm 0.56$ to $0.009 \pm 0.004$, a $99.4\%$ reduction.}
\label{fig2}
\end{figure}

\par For each trained seed, $\Phi_r$ was computed separately for $z$ and $g$ on the post-warmup portion of each rollout episode. As shown in Fig.~\ref{fig2}(a), the two distributions are clearly separated. Across $300$ trained episodes, mean $\Phi_r(g)=1.42$ $\pm$ SD $0.56$, whereas mean $\Phi_r(z)=0.011$ $\pm$ SD $0.021$ (paired $t=43.6$, $p<0.0001$). The inequality holds at the seed level for all $30$ trained seeds.

\par This localization, however, is partly predictable from the architecture itself. $g_t$ is a recurrent (GRU) latent, while $z_t$ is a feed-forward encoding, so a temporal information measure is expected to find more structure in $g$ than in $z$ on architectural grounds alone. The result is therefore best read as architectural localization, providing the anchor for the subsequent analyses: in this agent architecture, the $\Phi$ID-relevant signal is concentrated in the perspective latent $g$ rather than in the immediate perceptual representation $z$.

\par A second question is whether the large $\Phi_r(g)$ value reflects genuine temporal organization, or merely the marginal distribution of $g$. To test this, within-episode temporal shuffle was applied to each $g$ trajectory, preserving its empirical marginal distribution while destroying temporal order. As shown in Fig.~\ref{fig2}(b), this ablation collapses $\Phi_r(g)$ from $1.42 \pm 0.56$ to a near-zero residual of $0.009 \pm 0.004$, which marks a $99.4\%$ reduction (paired $t = 43.6$, $p<0.0001$). The signal therefore depends on the temporal ordering of the latent trajectory rather than on its static distribution.

\par These results establish the target of the remaining analyses. $\Phi_r$ in this architecture is concentrated in $g$, and the measured signal reflects temporal structure rather than a distributional artifact. The more substantive question is whether learning changes the $\Phi_r$ structure, and which atom-level components of $\Phi_r$ are most affected.

\subsection{Aggregate $\Phi_r(g)$ is supplied by architecture rather than learning}
\par The localization result shown in Section 3.1 does not by itself establish a learned effect. Because $g_t$ is a recurrent latent, high aggregate $\Phi_r(g)$ could arise from the gated recurrent substrate itself, even before training. To separate these contributions, I compare $\Phi_r(g)$ between trained and untrained agents under the same replay protocol.

\begin{figure}[t]
\centering
\includegraphics[width=0.5\textwidth]{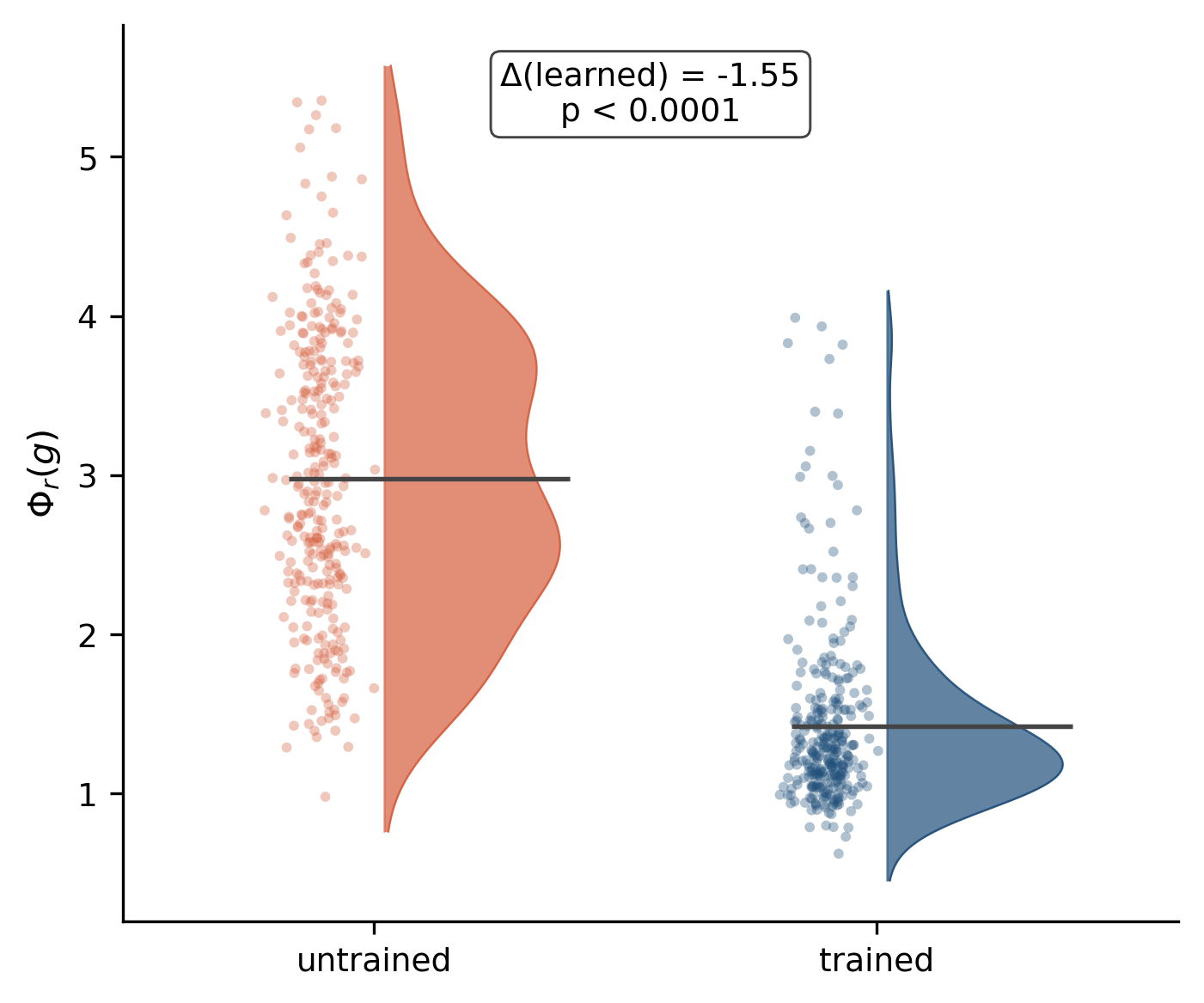}
\caption{\textbf{Aggregate $\Phi_r(g)$ is larger in untrained than trained agents.} Raincloud format follows Fig.~\ref{fig2}. Under the same replay condition, untrained agents show a higher mean $\Phi_r(g)$ value than trained agents (mean $\pm$ SD: $2.98 \pm 0.91$ vs.\ $1.42 \pm 0.56$).}
\label{fig3}
\end{figure}

\par The comparison highlights that the aggregate magnitude of $\Phi_r$ is not the operative quantity that tracks learning. Under the same replay condition, untrained agents show substantially higher aggregate $\Phi_r(g)$ than trained agents (Fig.~\ref{fig3}): $\Phi_r(g) = 2.98 \pm 0.91$ for untrained agents, compared with $1.42 \pm 0.56$ for trained agents (Welch's $t = -25.1$, $p < 0.0001$, Cohen's $d = -2.05$). Thus, the recurrent architecture already produces substantial $\Phi$ID-relevant temporal structure in $g$ prior to learning; learning lowers the scalar magnitude of this structure rather than increasing it. While~\cite{pigozzi2026causal} reports a direction-based alignment between $\Phi_r$ trajectories and reward improvement, the present cohort-level comparison shows that a simple magnitude-based reading of $\Phi_r$ would misattribute the substantive change of learning.

\par This result is important because it prevents aggregate $\Phi_r(g)$ from being interpreted as a direct measure of learned temporal organization. If scalar $\Phi_r$ were read in that way, the untrained agent would appear more integrated than the trained one. The more plausible interpretation is that aggregate magnitude conflates different atom-level contributions. The substantive effect of learning must therefore be sought not in the amount of $\Phi_r(g)$, but in its composition. This motivates the atom-group decomposition in the following section.

\subsection{Learning reorganizes the atom composition of $\Phi_r(g)$}
\par
\begin{figure}[t]
\centering
\includegraphics[width=\textwidth]{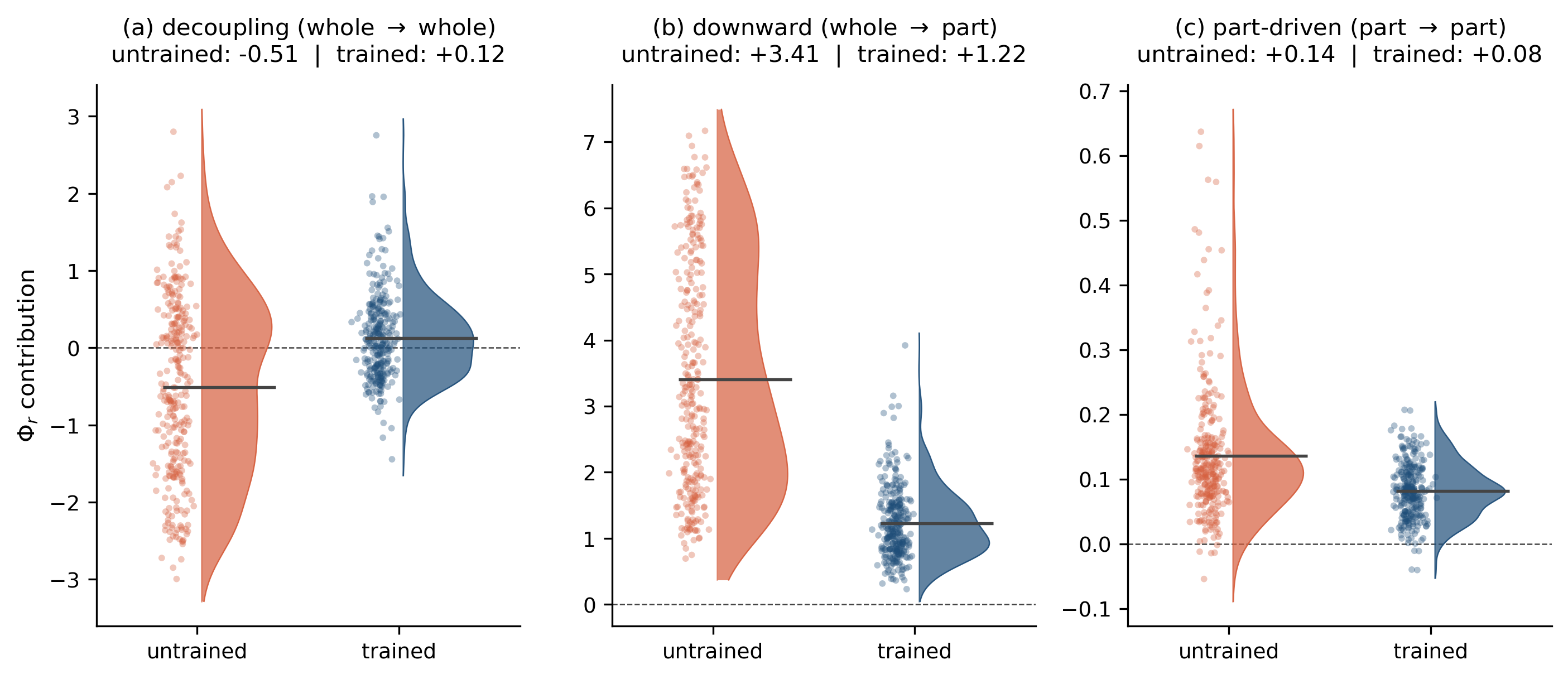}
\caption{\textbf{Learning reorganizes the atom composition of $\Phi_r(g)$.} Atom-group decomposition in untrained and trained agents under the same replay condition, with $300$ episodes per cohort. Dotted line indicates $y=0$. Cohort means are indicated as horizontal line, and also printed above each panel. \textbf{(a).} \textit{Decoupling} component shifts from negative in untrained agents to positive in trained agents. \textbf{(b).} \textit{Downward causation} component is reduced by roughly a factor of three after training. \textbf{(c).} \textit{Part-driven contribution} remain small in both cohorts.}
\label{fig4}
\end{figure}

 The preceding result shows that aggregate $\Phi_r(g)$ decreases after training. This raises the possibility that the scalar summary masks a change in the underlying compositional structure. To test this, $\Phi_r(g)$ is decomposed into the three atom groups defined in Section 2.4, and the resulting composition between trained and untrained agents is compared (Fig.~\ref{fig4}). Note that aggregate $\Phi_r(g)$ is computed as the median over time of the local sum of atoms, whereas the group means reported below are cohort means of episode-level atom values; the two therefore need not sum identically.

\par The \textit{decoupling} component changes in a distinctive way (Fig.~\ref{fig4}(a)). In untrained agents, decoupling is negative on average (mean $-0.51$), with positive seed-level means in only $9/30$ seeds. After training, the mean shifts to $+0.12$, with positive seed-level means in $22/30$ seeds ($\Delta_{\mathrm{trained-untrained}}=+0.64$, Welch's $t=8.6$, Cohen's $d=0.70$, $p<0.0001$). The change in absolute magnitude is smaller than for downward causation, but it is distinctive in sign shift. Positive decoupling indicates that the whole-to-whole atom contributes positively to the aggregate $\Phi_r$, whereas negative decoupling indicates an offsetting contribution. The shift from negative to positive decoupling therefore marks a qualitative change in the direction of the whole-to-whole contribution to $\Phi_r(g)$.

\par The largest change in magnitude occurs in the \textit{downward causation} component (Fig.~\ref{fig4}(b)). Untrained agents show a large downward contribution (mean $+3.41$), whereas trained agents show a substantially smaller value (mean $+1.22$; $\Delta_{\mathrm{trained-untrained}}=-2.18$, Welch's $t=-21.2$, Cohen's $d=-1.73$, $p<0.0001$). The high aggregate $\Phi_r(g)$ in untrained agents is therefore mainly carried by whole-to-part terms, and training reduces this dominant downward component.

\par The \textit{part-driven contribution} also shows a small but statistically significant decrease (Fig.~\ref{fig4}(c); $+0.14$ in untrained agents, $+0.08$ in trained agents; $\Delta_{\mathrm{trained-untrained}}=-0.05$, Welch's $t=-8.6$, Cohen's $d=-0.70$, $p<0.0001$). Its absolute magnitude is smaller than the other two groups in both cohorts, indicating that the architecture-versus-learning distinction is not primarily carried by bottom-up or lateral part-mediated terms.

\par The aggregate result in Section 3.2 can therefore be reinterpreted in compositional terms. Untrained recurrent dynamics produce high scalar $\Phi_r(g)$, dominated by downward whole-to-part contributions and accompanied by negative decoupling. Learning reduces the dominant downward causation and shifts decoupling positive, while leaving part-driven contributions small in both cohorts. In short, learning reorganizes the atom-level composition of $\Phi_r(g)$.

\subsection{Learned $\Phi_r(g)$ shows regime-specific atom dynamics}
\par The next question is whether the learned composition of $\Phi_r(g)$ remains stable under environmental change, or instead reorganizes when the observation-noise regime switches. Since prior work showed that $g$ exhibits hysteresis under regime change~\cite{pae2026aaai}, I ask here how the atom-level composition of $\Phi_r(g)$ responds to the regime switch. Throughout this section, $\Delta$ is reported as $\mathrm{post} - \mathrm{pre}$, while paired $t$-values are computed on the pre-vs-post difference, so their sign is the opposite of $\Delta$.

\begin{figure}[t]
\centering
\includegraphics[width=\textwidth]{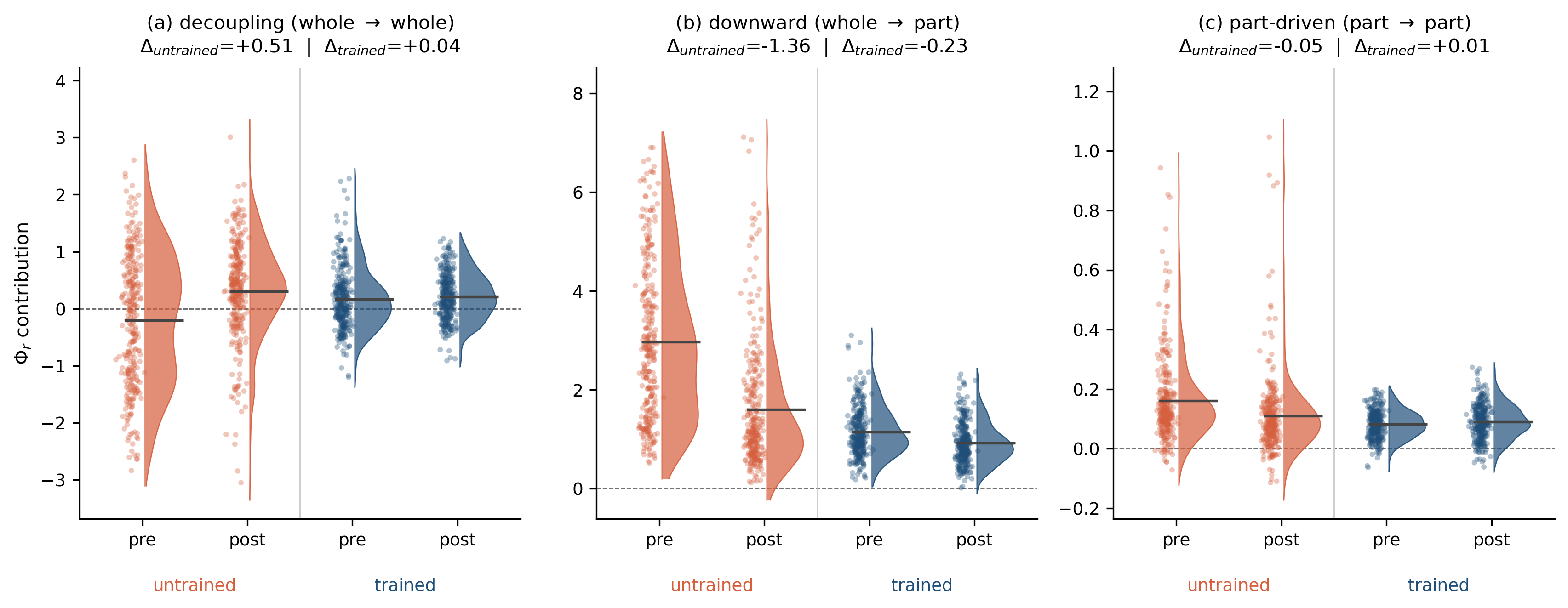}
\caption{\textbf{Atom-level pre-to-post regime switch composition.} Each panel shows one atom group, separated by cohort and pre/post switch window. Scatter points show individual episodes, half-violins show the pooled distributions, and horizontal bars mark means. In trained agents, decoupling and part-driven contributions remain approximately stable, while downward causation decreases. In untrained agents, all three groups shift across the regime switch, including a positive shift in decoupling and a large decrease in downward causation.}
\label{fig5}
\end{figure}

\par The atom-level decomposition clarifies what changes under the regime switch (Fig.~\ref{fig5}). In trained agents, the decoupling component is almost unchanged from pre- to post-switch ($\Delta=+0.04$, paired $t=-1.2$, $p=0.23$), and the part-driven component is likewise stable ($\Delta=+0.01$, paired $t=-1.9$, $p=0.06$). The main trained change is instead a decrease in downward causation ($\Delta=-0.23$, paired $t=6.6$, $p<0.0001$). Thus, the whole-to-whole component of $\Phi_r(g)$ remains stable across the regime switch, while the whole-to-part component adjusts to the new environmental regime.

\par Untrained agents show a broader and less selective response. Decoupling shifts upward from negative to positive values ($\Delta=+0.51$, paired $t=-7.0$, $p<0.0001$), downward causation decreases strongly ($\Delta=-1.36$, paired $t=14.2$, $p<0.0001$), and the part-driven component also changes significantly despite its smaller scale ($\Delta=-0.05$, paired $t=4.9$, $p<0.0001$). In other words, the untrained recurrent substrate has high aggregate $\Phi_r(g)$, but its atom-level organization is broadly regime-sensitive. By contrast, training confines the regime-dependent response mainly to downward causation while leaving decoupling approximately invariant.

\par This is the main mechanistic result of the analysis: learning does not create aggregate $\Phi_r(g)$; the recurrent architecture already supplies it. Instead, learning changes how $\Phi_r(g)$ is composed and how that composition responds to environmental change. In trained agents, the decoupling component shifts positive under learning and remains stable across the switch, whereas downward causation carries the main regime-dependent adjustment. These distinctions are invisible if $\Phi_r$ is treated only as a scalar magnitude.

\section{Discussion} \label{section4}
\par This study asked where $\Phi_r$-relevant temporal organization appears inside a reward-free active inference agent, and whether its scalar magnitude is sufficient to interpret the perspective latent $g$. The localization of $\Phi_r$ in $g$ is partly expected, since $g$ is the GRU-based latent whereas $z$ is a fast perceptual encoding. The more important finding is that the untrained baseline shows high aggregate $\Phi_r(g)$ can be supplied by the recurrent substrate alone. Scalar $\Phi_r$ therefore cannot be read as a straightforward index of learned perspective-like organization; it localizes a relevant temporal signal, but its meaning depends on how that signal is compositionally structured.

\par Learning reorganizes the composition of this temporal signal. In untrained agents, high aggregate $\Phi_r(g)$ is dominated by downward whole-to-part contributions and accompanied by negative decoupling. After training, aggregate magnitude decreases, but decoupling shifts positive and remains stable across regime change, while downward causation carries the main regime-dependent adjustment. In sum, learning does not create $\Phi_r(g)$ from nothing; instead, it transforms a high-magnitude recurrent signal into a more structured configuration in which whole-to-whole and whole-to-part components play distinct roles. Hence, the atom-level decomposition is essential for identifying what learning actually changes.

\par This connects to prior work that treats subjective perspective as a history-sensitive orientation rather than as a reward function or policy variable~\cite{pae2026aaai,pae2026sameworld,pae2026body}. In this framework, $g$ is introduced as a candidate structural locus through which the same environmental input is organized and perceived in history-dependent ways. The present analysis adds an information-dynamic layer to that proposal. It shows that learning gives $g$ a distinctive atom-compositional profile: the whole-to-whole component becomes positive and remains approximately invariant under regime shift, while downward causation carries the main regime-dependent adjustment. The atom-level decomposition therefore captures, in information-dynamic terms, what learning does to $g$ within this architecture.

\par This point is relevant for active inference more broadly. Active inference provides a principled account of perception, action, and policy selection without externally imposed reward~\cite{friston2017active,parr2022active,pezzulo2024sentient}. It is therefore natural to ask whether such reward-free predictive organization can also support perspective-relevant internal structure. In that context, the present results argue against interpreting $\Phi_r$ by magnitude alone. Prediction training produces temporally organized latent dynamics, but the relevant question is how those dynamics are compositionally organized. Specifically, $\Phi_r$ should be evaluated by whether it separates stable whole-level organization from adaptive whole-to-part engagement.

\par However, the interpretation should be kept deliberately modest. $\Phi_r$ is not presented here as a standalone measure of perspectival orientation in any phenomenologically rich sense. At most, the present results show that this quantity tracks a particular information-dynamic signature within a trained recurrent system. This follows from the minimal setting in which the analysis was conducted. The simulation uses a minimal gridworld. Moreover, the atom grouping is a topology-level decomposition of a reduced two-node system under Gaussian $\Phi$ID assumptions, not a complete description of all internal causal structure. Most importantly, the current design cannot separate effects of perspective-specific organization from effects of learned recurrence in general. Such distinction would require targeted architectural ablations, for example removing the stop-gradient separation so that $g$ is trained under actor-side objectives, or comparing against different recurrent latents that lack the perspective-specific coupling used here. Several of these limitations are the subject of ongoing work, which includes variance- and autocorrelation-matched baselines that isolate temporal structure from smoother recurrent dynamics, a partition-stability audit of the Fiedler bipartition across cohorts, an intermediate-checkpoint analysis of when the atom composition reorganizes during training, and a targeted stop-gradient ablation. 

\par Nevertheless, the central conclusion holds despite these limitations: aggregate $\Phi_r$ localizes the relevant temporal signal, while atom composition reveals what learning actually changes. For active inference and related approaches to perspective-relevant internal organization, this suggests that such structures should be sought in the compositional organization of temporally extended information flow.

\bibliographystyle{splncs04}
\bibliography{references}
\end{document}